# LEXSYS: Architecture and Implication for Intelligent Agent systems

**Robert ABC\*,**
\*LORIA 615, rue du Jardin Botanique, 54600 Villers les Nancy, France

**Abstracts**
*LEXSYS, (Legume Expert System) was a project conceived at IITA (International Institute of Tropical Agriculture) Ibadan Nigeria. It was initiated by the COMBS (Collaborative Group on Maize-Based Systems Research in the 1990. It was meant for a general framework for characterizing on-farm testing for technology design for sustainable cereal-based cropping system. LEXSYS is not a true expert system as the name would imply, but simply a user-friendly information system. This work is an attempt to give a formal representation of the existing system and then present areas where intelligent agent can be applied.*

**Keywords:** Agriculture, computer database, information system, intelligent agent, legume technology,

## 1.0 Introduction

LEXSYS, (Legume Expert System) was a project conceived at IITA (International Institute of Tropical Agriculture) Ibadan Nigeria. It was initiated by the COMBS (Collaborative Group on Maize-Based Systems Research in the 1990. It was meant for a general framework for characterizing on-farm testing for technology design for sustainable cereal-based cropping system.

The name LEXSYS, **L**egume **Ex**pert **Sys**tem was initially used though changes have being suggested. It is not a true expert system as the name would imply, but simply a user-friendly information system. The system has been updated severally despite the fact of its poor documentation. There had been several versions in DOS and the first window version came out in 1999. It had a remarkable uplift when the KBS version was developed by a research group at the department of Agricultural and Forestry Science, the University of Wales, United Kingdom under the project *"Education and research for a sustainable world"* . Two versions of LEXSYS-KBS have been developed.

LEXSYS lack proper documentation and this may suggest why follow-up is poor and localised. Results and contribution from users / researchers were communicated via email within the Agricultural research centres. The information dissemination is not wide and update is not regular. The aim of this work is to produce technical reference
- for anyone interested in the work
- for researchers/users who make use of it
- to develop a basis for integration into other similar systems
- promote cooperative information system development  and
- to develop a base for internet integration

## 2.0 Related works
Several works related to information system development has been done in the CGIAR centres. Some of these works are localised to a particular centre while others are inter-centre. LEXSYS was local and its main focus was IITA and other collaborating institutions. It was not developed on an existing framework of a particular system nor was it developed based on outside model. It was developed based on simple reflections on needs and prevailing problems associated with the integration of legumes into tropical farming systems at IITA and related NARS. The works reviewed here are generally works carried out in the CGIAR centres. They includes MPTS,

LIMS, SINGER and IRIS.

### 2.1 MPTS
The MPTS ( Multipurpose Tree and Shrub Database) was conceived and developed by ICRAF (International Centre for Research in Agroforestry). It is a user-friendly computing tool developed for the retrieval of detailed information on multipurpose tree and shrub species. The earlier versions of MPTS was a software that allows the user to browse species or create parameters for the selection of species based on any number of physical and social variables. The MPTS Users' manual describes it as, "a computer-based information and decision-support system founded on an inventory of existing information about multipurpose tree and shrub species in the tropics and sub-tropics. In the context of the database, multipurpose trees and shrubs are those that are deliberately grown or kept in integrated land-use systems and are often managed for more than one output.

The main objective of the database is to provide detailed information on a large number of species to field workers and researchers who are engaged in activities involving woody perennials suitable for agroforestry systems and technologies. It is designed to help them make rational decisions regarding the choice of candidate species for specific sites and defined purposes."

### 2.2 LIMS

**LIMS** (Laboratory Information Management Systems) is an information system being developed as joint collaboration between ICRISAT's scientists, Government of India, and Senior Scientist from the Directorate of Oilseeds Research all in Hyderabad, India. The system track laboratory samples, tests, results, archive data, and then produce timely and accurate reports reducing internal bottlenecks. It streamlines work and lowers costs while speeding processing times and turnaround. Computerized data collection and reporting systems built around a centralized database provides these productivity benefits. The Laboratory Information Management Systems, or LIMS, exist in many sizes and configurations, offering solutions to a host of applications and industries. Continuous work is going on on LIMS. Examples of new improvement being made include (a) sample tracking, (b) DNA extraction tracking, (c) DNA quantification tracking and (d) DNA dilution tracking. Newer version is hoped to be released for testing in first quarter of 2003. There is no indication of integrating this system to a wider internet forum.

### 2.3 SINGER
The System-wide Information Network for Genetic Resources (SINGER) is the genetic resources information exchange network of the International Agricultural Research Centres of the CGIAR. It provides access to information on the collections of genetic resources held by the CG centres. Together, these collections comprise over half a million samples of crop, forage and tree germplasm of major importance for food and agriculture.

It links the genetic resources information systems of the individual CG centers around the world, allowing them to be accessed and searched collectively.  SINGER contains key data on the identity, source, characteristics and transfers to users. It is a web-based system. Though SINGER makes rooms for contribution from users, feedback possibility is minimal.

Singer can be searched by
- **Taxonomy:** You can search for specific taxa based on the fields genus, species, crop and common names. This entry point allows you to identify specific taxa(s) and visualize detailed information on the collecting missions, accession level or material transfer areas.
- **Collecting missions**: This search is based on specific collecting missions carried by the CGIAR Centres and their co-operators by Centres, collections, taxa, country and year.

- **Accessions**: Search for can be based on accession level data. This entry points offers multiple choices (i.e. by CGIAR Centres, collections, taxa, country source, source of collection, sample status, ...).
- **Material transfer**: Seach can also be based material transfer or distribution. This entry point offers multiple choices from accession level data, cooperators name, organization, country, type and the date of transfer**.**
- **Cooperators**: You can search for cooperators that participated in collecting missions and/or in donated material and/or requested material.
- **Characterization & Evaluation**: You can search for characterization and evaluation data provided by the Centres.

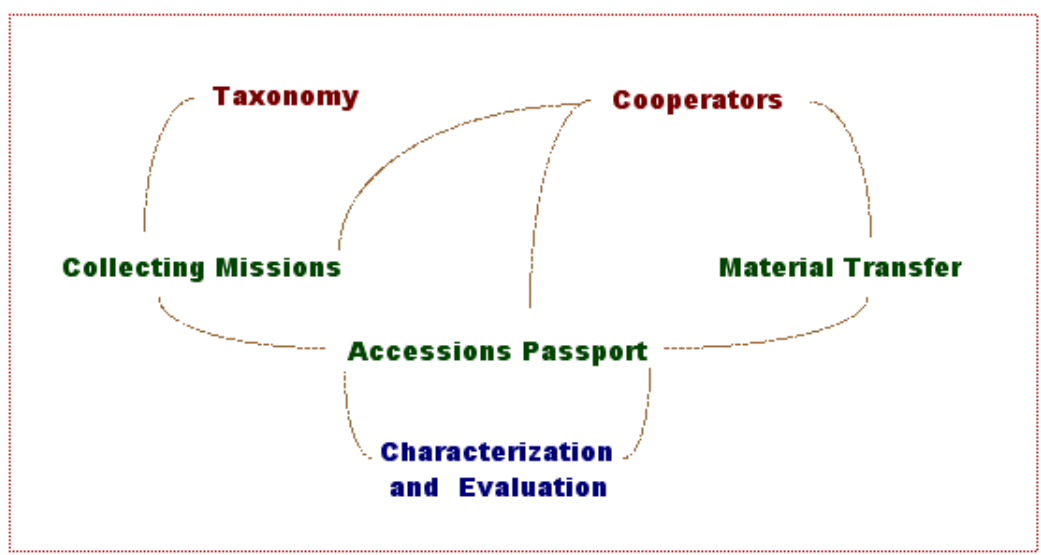

**Figure 1: Search entry points for SINGER (as presented on the Internet)**

**2.4 IRIS**
The International Rice Information System (IRIS) of the International Rice Research Institute (IRRI) is the rice implementation of The International Crop Information System (ICIS), a database system for the management and integration of global information on genetic resources and crop improvement for any crop. ICIS global project is not intended to be discussed here. Certain important aspect of ICIS will however be discussed. IRIS is being developed by genetic resource specialists, crop scientists and information technicians in several CGIAR centres and in National Agricultural Research Systems to solves problems related to

- ambiguous germplasm,
- difficulty in tracing germplasm pedigree information and lack of integration between genetic resources,
- characterization
- breeding,
- evaluation of germplasm
- intensive crop improvement programs.

The idea of ICIS was to provide a comprehensive crop information system that is implemented separately for each crop and is based on unique identification of germplasm, management of nomenclature (including homonyms and synonyms) and the retention of all pedigree information. The Genealogy Management System is the core database which performs these functions and links data from all disciplines in the crop improvement process. The distinct but compatible crop databases result in focused data management for each crop at the same time as benefiting from collaboration in the process of software development.

Collaborative software development is possible through adopting an open programming environment. Internationalization of the information systems is facilitated by a dual database design which gives remote users access to the central database through wide area networking or on compact disk. At the same time it allows users anywhere to add new pedigrees and data to a private, local database which can be exported periodically to the central database at the user's discretion.

The resulting crop systems will be invaluable for all researchers working on genetic resources or crop improvement, who will benefit from access to the latest international information linked unambiguously to specific germplasm. Just as these scientists have participated in the global exchange of genetic resources through the CGIAR's extensive networks, ICIS will allow them to benefit from, and participate in the development and deployment of new, knowledge-intensive, crop improvement systems that link information to the seeds or propagating material being exchanged

The problem associated with these systems are that in the case of LIMS, IRIS and MPTS the power of Internet is not harness to make the system robust enough for today modern information retrieval practices. In all of these systems reviewed (including SINGER), user's need, interest and habits were not considered.

### 3.0 Overview of LEXSYS
LEXSYS was initially developed with top-down programming method. The KBS version approached the development using WIN-PROLOG. The KBS version of LEXSYS is presented in Figure 2 and the general concept and program flow is presented Figure 3.

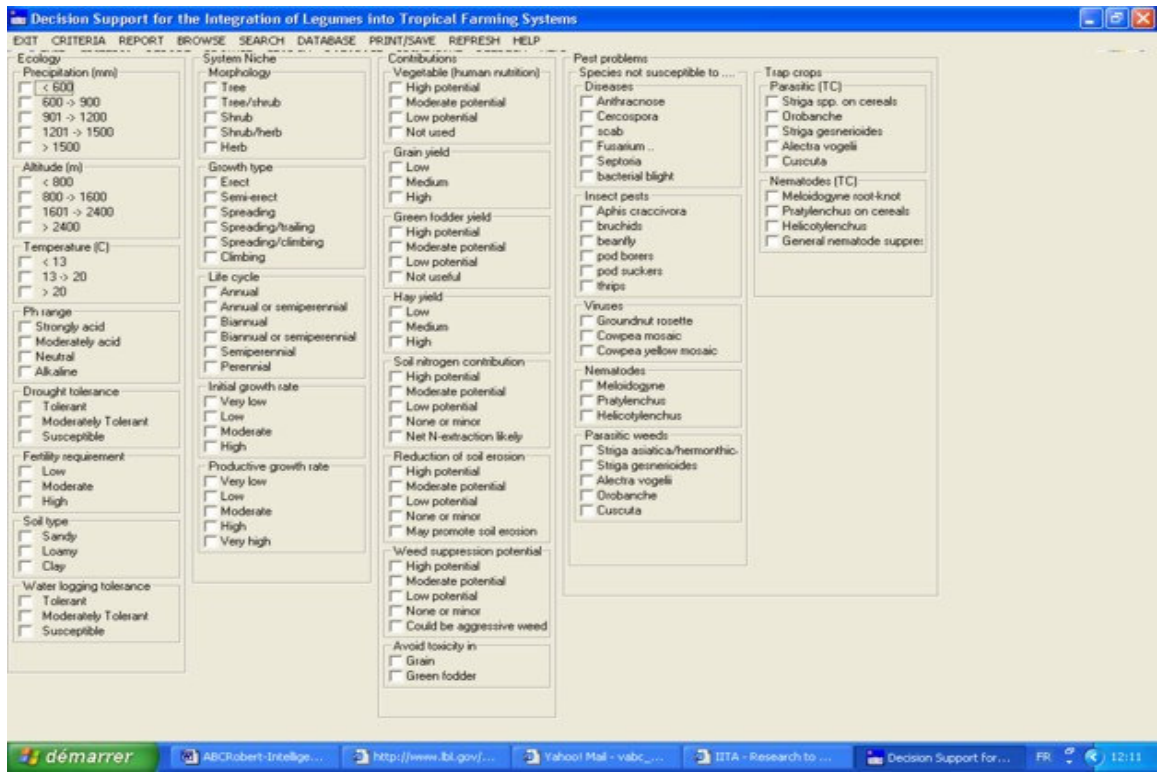

**Figure 2: Interface of LEXSYS (KBS)**

In general, at the core of the program, it passes defined parameters and their respective values into a global decision subroutine. The selection criteria are stored in a file for reference purpose. Only species whose characteristics match those of the criteria are selected and stored in a file for future purposes. WHY, COMBINE and BROWSE subroutines reference these stored information. The parameters for selection are categorised into five broad groups. These groups were further sub-divided into sub-groups. The groups and their respective sub-groups were defined by scientists at IITA, NARS and other international research organisations.

A metal-model proposed in LORIA[1] by the SITE research group will be adopted in describing the parameters and values used in the selection process. The format of this model is Property(options). It should be made clear that definition of properties and values are defined based on (a) field experiences and (b) prevailing conditions in the tropical and sub-tropical regions.

---

[1] **LORIA**:Laboratoire Lorrain de Recherche en Informatique et ses applications (www.loria.fr)
**SITE** : Modélisation et Développement de Systèmes d'Intelligence Économique

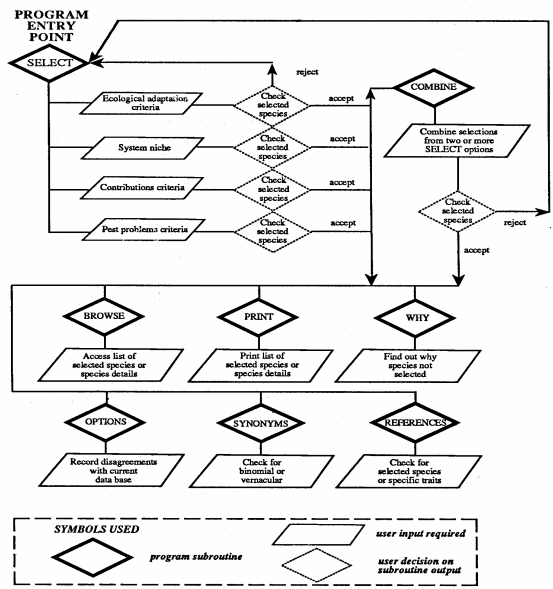

**Figure 3: LEXSYS 2.0 Program flow**

```
{Select}
    {Ecology}
        {Precipitation(<601|601-900|901-1200|1201-1500|> 1500)}
        {Altitude range(<800|801-1600|1601-2400|> 2400)}
        {Temperature range(<12|13-20|> 20)}
        {Drought risk(Low risk|Short drought|Moderate drought|Extended drought )}
        {Ph-range( Strongly acid | Moderately acid |Neutral|Alkaline )}
        {Fertility range(Very low|Low|Moderate)}
        {Soil texture(Sandy|Loamy|Clay)}
        {Water logging(None|Temporary|Seasonal|Permanent)}
   {System niche}
        {Morphology( Tree/shrub|Shrub|Shrub/herb|Herb|Any one )
        {Growth type(Erect|Erect/spreading|Spread/non climbing
                |Spreading/climbing|Climbing | Any one)}
        {Life cycle(Annual |Annual/semi-perennial|Semi-perennial/perennial
                |Perennial|Any one)}
        {Initial growth(Low|Low/moderate|Moderate|Any one)}
        {Productive growth(Moderate|Moderate/high|High |Any one')}
   {USE}
        {Vegetable for human nutrition(Low/moderate|Moderate/high
                |Avoid highly toxic|Not relevant)}
        {Grain for human nutrition(Low/moderate grain yield|Moderate/high grain yield
                |High grain yield|Avoid highly toxic grain'|Not relevant'
        {Fodder (green)(Low/moderate potential|Moderate/high potential
                |Avoid highly toxic plants|Not relevant)}
        {Hay for livestock(Low/moderate yield|Moderate/high yield
                |High yield|Avoid toxic residues|Not relevant)}
        {Soil improvement(Soil nitrogen|Soil erosion|Weed suppression)}
   {Trap Parasites}
        {Parasitic weeds(Striga hermonthica/asiatica|Striga gesnerioides'
                |Alectra vogelii |Orobanche spp.| Cuscuta spp.)}
        {Nematodes(Pratylenchus|Meloidogyne)
   {Avoid Susceptibility}
        {Diseases(Anthracnose|Bacterial Blight|Scab |Septoria |Fusarium Wilt)}
        {Insect pests(Beanfly|Aphis craccivora|Flower thrips|Pod Sucking Bugs)}
        {Viruses(Groundnut Rosette |Cowpea Mosaic|Cowpea Yellow Mosaic)}
        {Nematodes(Meloidogyne|Pratylenchus |Helicotylenchus)}
        {Parasitic weeds(Striga hermonthica/asiatica|Striga gesnerioides'
                |Alectra vogelii |Orobanche spp.| Cuscuta spp.)}
{/Select}
```

## 3.1 Internal Architecture

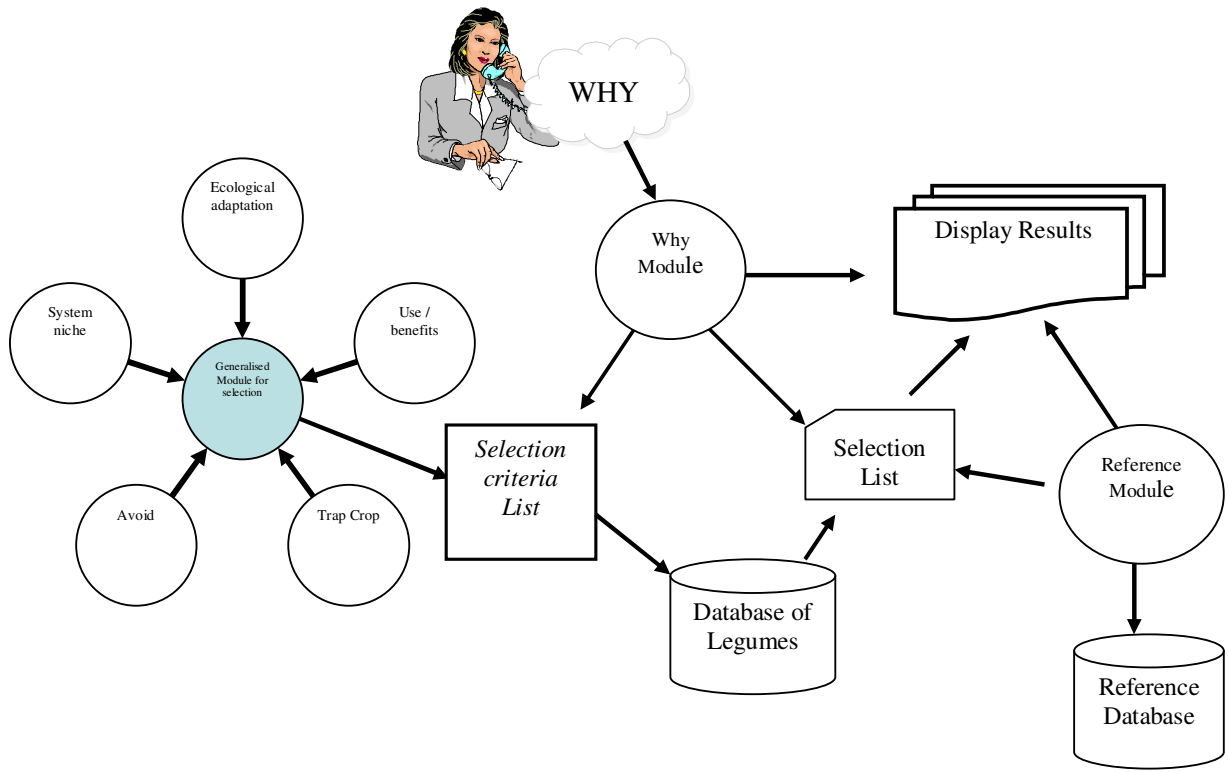

### 3.1.1. Selection
The core of LEXSYS is the central query modules which takes inputs from one or more of the 47 search criteria classified into five groups (i.e. ecology, system niche, uses/contribution, trap crop and avoid).

### 3.1.2 Ecology
In ecology, the selection criteria are based on climatological parameters such as rainfall, altitude, mean temperature, soil characteristics, drought and water logging. The selected parameters and values are primarily the interest of the user or prevailing situations in the geographical zone, environment or ecology of his/her choice.

### 3.1.3 System Niche
The system niche identifies plants' biological characteristics of legume species that can be adapted to an existing farming system. These characteristics are morphologic, physiologic or phenologic

### 3.1.4 Contribution/Use
In contribution the attention is application of plant products or plant characteristics for one or more human needs. The need could be industrial, domestic or economic.

### 3.1.5 Pest-Avoid
In Pest-Avoid selection, the concern is to select plants that do not have one of more of the traits identified as parameter in the selection.

### 3.1.6 Pest-Trap
Pest-trap identifies crops that can be used to trap or combat one or more plant problems. For example, in recent years, *Striga sp*, has been noted to be one of the main hindrances to maize productions in the tropics. LEXSYS may be used to find out a legume that can combat this menace.

### 3.2 Selection module
The method adopted is to save criteria list in a predefined array. This is saved as arguments type definition and its correspondent value for in arrays. These arrays are evaluated from the database of legume. The typical programming example is shown below.

```
{Evaluate}
(VAL(&q[1])<u[1].AND. &q[1]#' ').AND. (VAL(&r[1])>v[1] .OR.;
    &r[1]=' ') .AND. (VAL(&q[2])<u[2].AND. &q[2]#' ').AND. (VAL(&r[2])>v[2] .OR.;
    &r[2]=' ') .AND. (VAL(&q[3])<u[3].AND. &q[3]#' ').AND. (VAL(&r[3])>v[3] .OR.;
    &r[3]=' ')
……….

{/Evaluate}
```
where $q(n)$ and $r(n)$ are the argument types while $u(n)$ and $v(n)$ are the argument values

### 3.3 Why Module
The why module is also central in LEXSYS because it gives feedback room to check why a pre-conceived legume species is not included in a desired selection. This module makes use of the selection list, criteria list and the database (which contain the selection property type and value). Typical programming example is

```
Procedure Why(Parameter(Legume Species in mind))
{Evaluate}
goto <legume species>
If ((VAL(&q[k])<u[k] .AND. &q#' ') .AND. (VAL(&r[k])>v[k] .OR. &r[k]=' '))
        then
    (q[k])=[critType],'Not adapted to &[critType]
end if
{/Evaluate}
```

where $q[k]$ is the lower bound $k^{th}$ selection criteria type and $u(k)$ is the lower bound $k^{th}$ selection value. $r[k]$ is the $k^{th}$ upper bound property type in the legume database and $v[k]$ is the $k^{th}$ upper bound property value in the database

### 4.0 The Intelligent Agents
One definition of intelligent agents is that they are systems aim at helping the user in finding information. This is not the only application of intelligent agent but LEXSYS can be developed bearing this piece of information in mind. One of the widely method adapted in intelligent agents is the machine learning. Two approaches are used in incorporating machine learning into intelligent agents. These are content-based and collaborative approaches [Mladenic 99].

In the content-based approach, user's request is matched with the content of WEB sites using one of the more than eight different algorithms [Lewis et al ]. Whatever model or algorithm that is used is not our immediate concern in this paper. In the collaborative approach, user's profile or information being researched is complemented with inputs from other users who have similar profile or closely-related profile or who had made similar research.

## 4.1 Application of Intelligent Agents

The prediction 'Agent-based computing (ABC) is likely to be the next significant breakthrough in software development' (Sargent, 1992), can be well said to be true today considering the present methods being used in software development. We can adapt newer developments in LEXSYS in this direction. It is believed that agency approach to its development will give it a significant uplift. First, the use of LEXSYS requires agronomic, economic and farming knowledge (Weber G.K et al). This implies that it requires substantial amount of inputs from the user and researcher from the field. Secondly, it is user-dependent. This implies that user's profile can be of significant importance in its relevancy. Previous versions generally do not reference user's profile when selection is made on leguminous crops for any of the five categories of selection. Every run is a fresh request for user's intentions. At the close of program, when the program is terminated, the system removes all the temporary files that were used in making decisions including selection lists. This implies that (a) the cumulative attitude of the user is not considered while executing the program (b) user's interest can not evolve. An intelligent agent if integrated to this system will enhance its usefulness in several ways. Among these are:

- It should provide possible framework for system performance analysis
- It should provide way of accessing improvements to be made in future version
- It should improve "run time" access

Three ways of incorporating agents into an application was suggested by Mohammadian M. (Mohammadian, 99). (a) The agents is integrated as part of the application (b) The agent is separated from the application and it has extensive domain specific information about the application (c) The agent can learn the user's preferences based on his responses. Incorporating intelligent agent will not only manage the user's profile but assist in rapid information retrieval.

Intelligent agent application in LEXSYS is being proposed because of the following properties of intelligent agents

- *Taskable*. The intelligent agent can take task and command from human or other agents
- *Semi-autonomous*. Intelligent agents can act on their own without intervention from humans
- *Persistent*. Activities of intelligent agents can take place for long period without attention
- *Reliable*. Intelligent agent can repeat the same activity with the same level of accuracy over and over again
- *Active*. An agent should anticipate and act appropriately on behalf of users
- *Collaborative*. Agent can collaborate with human or with other agents. This enables it to increases their local knowledge, resolve problems and inconsistencies. In effect improves its decision support capabilities.
- *Adaptive*. It adapt to changing user's need and task environment.

LEXSYS can be considered as a simple information retrieval systems without a feedback possibility. The power of LEXSYS should be extended if some or all of the following areas are addressed

- Databases
    - Legume Database
      There are just 123 items in the legume database. The list of legumes in the database is not exclusive. With possible inclusion of collaborative agents, researchers in the fields can of course add to or subtract from the list. An attempt

was made by introducing an option called "note". The "note" option was supposed to gather information from the researchers on the field.
- Reference database
  Reference database includes list of some six hundred bibliographic citations on legume characterization. With the possibility of intelligent agents, this list can be improved.
- Program development
  - Selection criteria
    Selection criteria can be improved by (a) contribution from researchers in the field or (b) from users' profile
  - Why
    Through the use of intelligent agent "Why option" could suggest preference of legume species. This can be included in program development. For instance, a pull down menu of most referenced species can be built-in during program development.

- Usage and recommendations
  - since most researchers/users will not make time to respond to questionnaires, an intelligent agent could suggest the usage pattern and the value attributed to the software.

- In general
  - there should be possibility to add an online feedback from the user
  - two or more users could interact/collaborate online
  - the database should be dynamic: individual user should be able to fine-tune their database to their own use either by creating a local copy of database which is a subset of the central database . They should be able to also contribute to the central database
  - individual user's habit should be integrated and
  - synonyms of definitions could be suggested by users.
  - online documentation and reference should evolve through users' participation

**5.0 Suggested modules/areas for Intelligent Agent**
The area where intelligent agent can be applied includes:

**5.1 Selection**
- **Pattern:** Various users have a pattern (unintentional and unknown to them) accessing and using the software. This pattern is not made available to the system developer. If this pattern is included in a feedback, it should improve the performance of the system
- **Method:** The method of selection should also be made to vary according to individual user. This could also be used to enhance system performance
- **Criteria:** Criteria for selection are believed to be too rigid and can not be said to represent general needs.
- **List:** The list for selection and selected species list should also contribute to future selection

**5.2 Why**
Why procedure should be intelligent in that request made is referenced to (a) previous similar request (b) initiating selection criteria in the past (c) previous similar request and initiating selection criteria combined together

**5.3 Referencing module**
It will be impossible for one group of localised individuals to have exhaustive list of publication that can be used in characterizing the criteria or describing the characteristics of individual species properties. The internet can be used to gather broad-based, multi-lingua, multi-regional and all-encompassing references. In a nutshell, an intelligent agent will improve system referencing.

**6.0 Conclusion and recommendations**
It is believed that this work should provide (particularly for LEXSYS) insight for improved system development based on
- the power of internet
- collaborative system development facilities
- the power of distributed information gathering
- the power of intelligent agency

This paper can be a launch pad for future development not only in LEXSYS but for projects in information system development or even in agricultural projects. It can also be a motivation for greater collaborative work in general.

**7.0 Future Possible Research**
Most of the contributing authors/researcher involved in the development of LEXYS is biased. Their field of research is related to agriculture, they may not provide accepted view for a generalised audience. It will be interesting if LEXSYS can incorporate other disciplines such as pharmacy, medicine, chemistry and biology. This will greatly enhance the its credibility and the usability.

## Internet Sites

ICRAF's Multipurpose Tree and Shrub Database, http://www.ciesin.org/IC/icraf/mptsdata.html

Harnessing biotechnology for the poor, http://www.icrisat.org/text/research/grep/bioinformatics/dfid/dfid.htm
The CGIAR System-wide Information Network for Genetic Resources,
    http://singer.cgiar.org/Search/SINGER/search.htm
The international rice information system, http://www.icis.cgiar.org/ICIS_IRRI.htm
The University of Wales, http://www.safs.bangor.ac.uk/
LORIA**:** http://www.loria.fr
SITE: http://www.loria.fr/LORIA/PHP/equipes.php?menu=equipesloria&organ=loria&langue=fr&fichier=Site

## ACRONYMS
CGIAR or CG: Consultative Groups on International Agricultural Research (www.cgiar.org)
COMBS: Collaborative Group on Maize-Based Systems Research
ICIS: International Crop Information System
ICRAF: International Centre for Research in Agroforestry , Nairobi, Kenya (www.icraf.org)
ICRISAT: International Crop Research Institute in the Semi-Arid Tropics, Patancheru, India (www.icrisat.org)
IITA: International Institute of Tropical Agriculture, Ibadan, Nigeria (www.iita.org)
ILRI: International Livestock Research Institute, Adis Ababa, Ethiopia (www.ilri.org)
IRRI: International Rice Research Institute (www.irri.org)
KBS: Knowledge-based system
LEXSYS: Legume Expert system
LORIA:Laboratoire Lorrain de  Recherche en Informatique et ses applications
NARS: National Agricultural Research Systems
SITE : Modélisation et Développement de Systèmes d'Intelligence Économique

## LEXSYS Site
   ftp://ftp.bangor.ac.uk/pub/departments/af/LEXSYS/